\pdfoutput=1
\documentclass[11pt]{article}

\usepackage[]{acl}

\usepackage{times}
\usepackage{latexsym}

\usepackage[T1]{fontenc}
\usepackage[utf8]{inputenc}

\usepackage{microtype}

\usepackage{inconsolata}

\usepackage{url}            % simple URL typesetting
\usepackage{amsmath,amsthm,amsfonts,amssymb,bm,stmaryrd}       % blackboard math symbols
\usepackage{nicefrac}       % compact symbols for 1/2, etc.
\usepackage{enumitem}
\usepackage[noend]{algpseudocode}
\usepackage{algorithm}
\usepackage{mathtools}
\usepackage{nccmath}
\usepackage{multirow}
\usepackage{bigdelim}
\usepackage{color, colortbl}
\usepackage{xcolor}		% make links dark blue
\definecolor{darkblue}{rgb}{0, 0, 0.5}
\hypersetup{colorlinks=true, citecolor=darkblue, linkcolor=darkblue, urlcolor=darkblue}
\usepackage{booktabs}
\usepackage{dcolumn}
\newcolumntype{d}{D{.}{.}{-1}}
\newcommand\hd[1]{\multicolumn{1}{c}{\textbf{#1}}}
\usepackage{graphicx}
\usepackage{titlesec}
\usepackage{hyperref}
\usepackage{url}
\usepackage{framed}
\usepackage[most]{tcolorbox}
\usepackage{microtype}
\usepackage{subcaption}
\usepackage{graphicx}
\usepackage[normalem]{ulem}
\usepackage{xspace}
\useunder{\uline}{\ul}{}
\usepackage{fontawesome}
\usepackage{calc}
\newcommand\mytokens[1]{\mytokenshelp#1 \relax\relax}
\def\mytokenshelp#1 #2\relax{\allowbreak\grayspace\tokenscolor{#1}\ifx\relax#2\else
 \mytokenshelp#2\relax\fi}
\newcommand\tokenscolor[1]{\colorbox{gray!10}{\textcolor{black}{%
  \ttfamily\mystrut\smash{\detokenize{#1}}}}}
\def\mystrut{\rule[\dimexpr-\dp\strutbox+\fboxsep]{0pt}{%
 \dimexpr\normalbaselineskip-2\fboxsep}}
\def\grayspace{\hspace{0pt minus \fboxsep}}

\title{SEMQA: Semi-Extractive Multi-Source Question Answering}

\author{Tal Schuster,$^{1}$ \quad Adam D. Lelkes,$^{1}$ \quad Haitian Sun,$^{2}$ \quad Jai Gupta,$^{1}$  \\ \\ \textbf{Jonathan Berant,$^{2}$ \quad William W. Cohen,$^{2}$ \quad Donald Metzler$^{1}$}
\\
\\
$^1$Google Research \qquad $^2$Google DeepMind %\\
}

\definecolor{hl_1}{HTML}{f79d9c}
\definecolor{hl_2}{HTML}{9df7c4}
\definecolor{hl_3}{HTML}{cfdfff}
\definecolor{hl_4}{HTML}{fcf77e}
\definecolor{hl_5}{HTML}{ffe5cc}
\definecolor{hl_6}{HTML}{c8c8b5}
\definecolor{hl_7}{HTML}{def957}
\tcbset{on line, 
        boxsep=1pt, left=0pt,right=0pt,top=0pt,bottom=0pt,
        colframe=white,colback=hl_1,  
        highlight math style={enhanced}
        }

\newcommand{\thedataset}{QuoteSum\xspace}

\newcommand{\parhead}[1]{\paragraph{#1}}

\begin{document}
\maketitle
\begin{abstract}
Recently proposed long-form question answering (QA) systems, supported by large language models (LLMs), have shown promising capabilities. Yet, attributing and verifying their generated abstractive answers can be difficult, and automatically evaluating their accuracy remains an ongoing challenge.
In this work, we introduce a new QA task for answering multi-answer questions by summarizing multiple diverse sources in a semi-extractive fashion. Specifically, Semi-extractive Multi-source QA (SEMQA) requires models to output a comprehensive answer, while mixing factual quoted spans---copied verbatim from given input sources---and non-factual free-text connectors that glue these spans together into a single cohesive passage. This setting bridges the gap between the outputs of well-grounded but constrained extractive QA systems and more fluent but harder to attribute fully abstractive answers. Particularly, it enables a new mode for language models that leverages their advanced language generation capabilities, while also producing fine in-line attributions by-design that are easy to verify, interpret, and evaluate. To study this task, we create the first dataset of this kind, QuoteSum, with human-written semi-extractive answers to natural and generated questions, and define text-based evaluation metrics. Experimenting with several LLMs in various settings, we find this task to be surprisingly challenging, demonstrating the importance of QuoteSum for developing and studying such consolidation capabilities.\footnote{Dataset and code are available at: \url{https://github.com/google-research-datasets/QuoteSum}}\looseness=-1

\end{abstract}

\section{Introduction}\label{sec:intro}
Large Language Models (LLMs) have recently demonstrated impressive capabilities across NLP tasks~\citep{srivastava2022imitation} and in particular, in answering general questions~\citep{gpt4, palm2}. As a result, an increasing number of users interact with such models when seeking information. Although LLMs often return high-quality and correct results, they can still make convincing-sounding mistakes (e.g., relying on false or out-dated information~\citep{dhingra-etal-2022-time,schuster-etal-2020-limitations,schuster-etal-2021-get}). Added citations can help the reader verify the generated answers~\citep{bohnet2023attributed,Gao2023EnablingLL}, but might be inaccurate themselves and add a false impression of reliability~\citep{Liu2023EvaluatingVI,Yue2023AutomaticEO}.
Furthermore, evaluating fully-abstrative long-form QA systems remains a challenge, mostly involving either running another large QA model or expensive human ratings.

\begin{figure*}[t]
    \centering
    \includegraphics[width=0.95\textwidth]{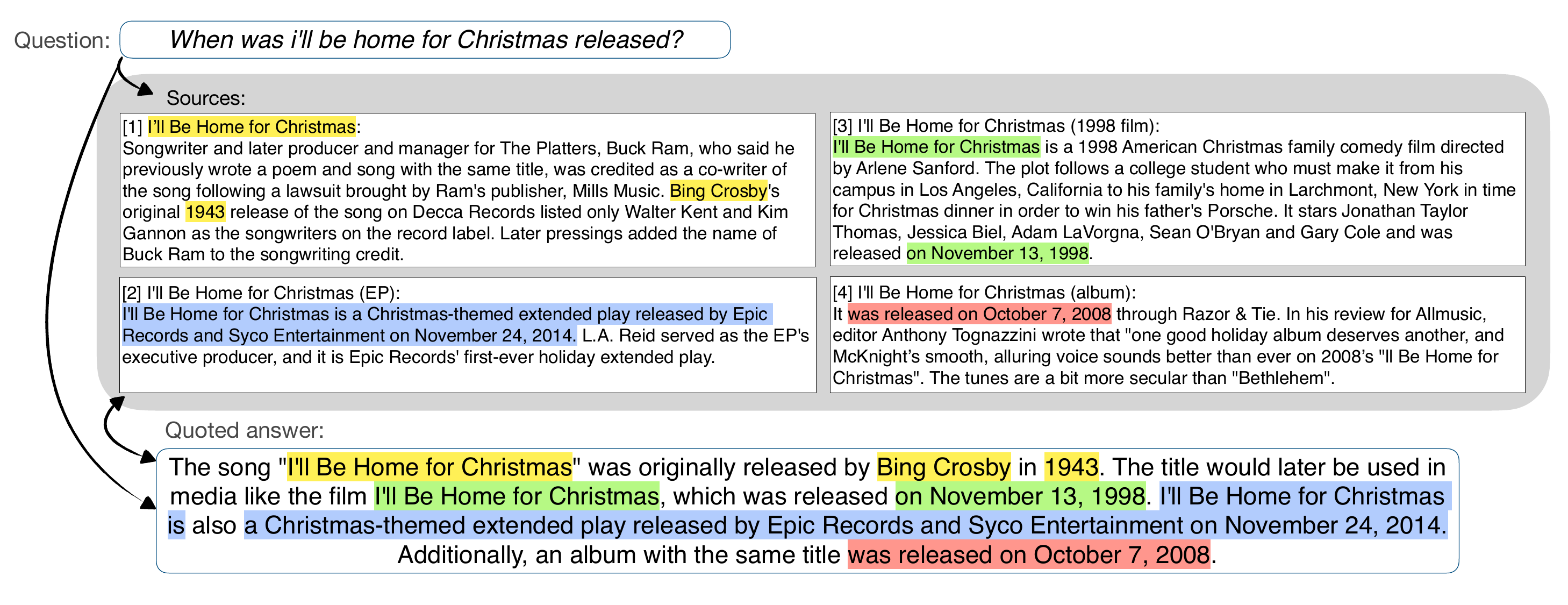}
    \vspace{-8pt}
    \caption{In SEMQA, the answer combines information from multiple sources while explicitly extracting factual spans (highlighted in different colors per source), and connecting them into a coherent well-grounded passage.}
    \label{fig:intro}
    \vspace{-5pt}
\end{figure*}

% our task
In this work, we introduce the task of semi-extractive multi-source QA (SEMQA) for answering multi-answer questions. Specifically, given a question and a set of retrieved passages, the goal is to generate a summarized and well-grounded answer that interleaves verbatim extracted spans of factual statements with free-text connectors.

% benefits of our task 1. grounding. 2. evaluation. 3. speed 
Including explicitly extracted spans in long generated answers provides multiple advantages. For example, it readily supports fast and easy verification by the reader---only requiring to confirm the context in the source. Also, it simplifies and standardizes the evaluation process. Instead of relying on model-based metrics that are expensive to run, hard to interpret, and depend on the performance of the measuring model~\citep{deutsch-etal-2022-limitations}, extracted spans can be readily evaluated for both recall and precision with standard string-matching metrics~\citep{rajpurkar-etal-2016-squad}. Finally, explicitly quoting spans can simplify the expensive auto-regressive generation process of LMs, enabling potential efficiency gains~\citep{mallinson-etal-2022-edit5,schuster2022confident}.\looseness=-1

Consider the example in Figure~\ref{fig:intro} that asks when was \emph{``I'll be home for Christmas''} released. A fully-extractive format would likely only return the set of short-answers without any contextualization (e.g., \emph{1943; November 24, 2014;...}). Instead, in SEMQA, dates, relevant entity names, and factual statements are extracted and glued together into a cohesive and well-grounded passage.  

% how we collect the data
To study the SEMQA task, we construct a dataset of questions, relevant passages, and human-written semi-extractive answers. Our dataset, named QuoteSum, includes both natural multi-source questions~\citep{kwiatkowski-etal-2019-natural,min-etal-2020-ambigqa} and generated multi-answer questions~\citep{lewis-etal-2021-paq}. We created an annotation pipeline where writers are asked to compile answers while using a tool that visualizes and keeps track of copied spans. In the task definition, we ask writers to use the provided quoting mechanism for all factual statements, while ensuring fluency of the text. Writers can choose to skip certain sources if they don't include additional helpful content and cite only a subset of the given sources, simulating possible noise in the retriever module (see \S\ref{sec:dataset}).

We choose to focus on the multi-source setting since we argue it to more clearly benefit from this style. However, our approach can be easily relaxed to single-source and single-answer questions. By focusing on multi-answer questions we also emphasize the main challenges of this task, requiring models to (1) correctly find an answer in each source; (2) extract relevant context from each source to contextualize the short answers; and (3) effectively consolidate the different answers according to their relations with each other and to the question.

% models we try on
We use \thedataset to fine-tune large QA models and evaluate both supervised and in-context LLMs. We find LLMs to obtain remarkable performance with only a few \thedataset examples. However, \thedataset-tuned models perform best, indicating the usefulness of our data. We also run a user study comparing SEMQA answers to cited abstractive answers~\citep{Gao2023EnablingLL}, and find SEMQA answers to be more comprehensive and much easier to be manually verified by readers.

Overall, our experiments reveal that the SEMQA task is surprisingly challenging even for modern LLMs. We define text-based metrics for SEMQA, creating easy-to-compute model-free measures to track the progress of models towards solving this task, capturing the fluency, preciseness and comprehensiveness of the generated quoted answers. 

% main contributions
Our main contributions include:
\begin{itemize}[leftmargin=*,noitemsep]\vspace{-3pt}
    \item Introducing and formulating the task of semi-extractive multi-source QA (SEMQA).\vspace{3pt}
    \item Creating \thedataset, the first dataset for this task, including multi-answer questions with high-quality human-written semi-extractive answers.\vspace{3pt}
    \item Evaluating different LLMs using both text-based metrics and via a user study, revealing SEMQA's challenges and facilitating future research.
\end{itemize}

\section{Related Work}\label{sec:related}
Several open-domain QA datasets use short extracted spans or entities as target answers~\citep{berant-etal-2013-semantic,joshi-etal-2017-triviaqa,kwiatkowski-etal-2019-natural,rajpurkar-etal-2016-squad,rajpurkar-etal-2018-know}.
The recent advancements in LLMs allowed expanding automatic QA systems to more challenging settings such as long-form answers~\citep{eli5_lfqa}, answers with multiple entities~\citep{amouyal2023qampari,zhong2022romqa}, and ambiguous questions~\citep{min-etal-2020-ambigqa,stelmakh-etal-2022-asqa}.
While most recent LLM-based QA models focus on abstractive free-text closed-book or retrieve-then-read settings~\citep{bohnet2023attributed,NEURIPS2020_6b493230}, evaluating these models remains a challenge~\citep{Min2023FActScoreFA}, even when answers have citations~\citep{Gao2023EnablingLL,Huang2023CitationAK, hagrid,Li2023TowardsVG, Liu2023EvaluatingVI,Malaviya2023ExpertQAEQ, Yue2023AutomaticEO}---mostly involving expensive model-based measures~\citep{honovich-etal-2022-true-evaluating,rashkin2022measuring}.

Similar to our motivation to improve the faithfulness of generated answers, \citet{potluri2023concise} suggest an extract-and-decontextualize pipeline that starts with a fully extractive long-form answer, and edits it for improving fluency. However, they focus on single-source questions which don't require multi-answer consolidation.

Regarding evaluation, recent studies identified some limitations of lexical metrics for general QA evaluation~\citep{bulian-etal-2022-tomayto,wang2023evaluating}. However, our task formulation that limits the output space to the given sources, combined with the semi-extractive nature and collecting multiple references, allow us to define model-free evaluation metrics that overcome most of the discussed limitations (see \S\ref{sec:task_def}). Ultimately, a practitioner might choose to combine multiple different evaluators.

\section{Semi-Extractive Multi-Source QA}\label{sec:task_def}
To formally define the SEMQA setting, we assume an input question $q$ and a set of two or more passages $\mathcal{P}$ that possibly include an answer to question $q$. To simplify the setting, we assume here that the passages $p\in\mathcal{P}$ are already selected by some retriever that models $\mathbb{P}(p \mid q)$. Our goal is to generate an answer $(q, \mathcal{P}) \rightarrow a$ that (1) answers $q$, covering all components and aspects based on information from $\mathcal{P}$, (2) is as extractive as possible with respect to $\mathcal{P}$ (explicitly marking extractions and their source), and (3) is concise\footnote{Our ``concisenss'' objective is within the space of semi-extractive answers. Naturally, unconstrained fully-abstractive answers can be more concise, but don't satisfy objective (2).}  and fluent.

There is a natural trade-off between the last two requirements for $a$. This balance is necessary since naïvely  maximizing only (1) and (2) would lead to simply concatenating all passages in $\mathcal{P}$. In contrast, maximizing only (3) would lead to an empty answer. We resolve this tension by seeking to ensure extractiveness for entities and for core factual statements, and to prefer fluency otherwise.

By explicit quoting marks, we mean that each extracted span should be marked as such, with an indicator for which passage it was taken from. The passage indicator is important for distinguishing identical spans across passages (e.g., \emph{``I'll be home for Christmas''} in Figure~\ref{fig:intro}) and connecting each statement in the answer with the original supporting source. For brevity, we refer to such explicitly marked semi-extractive answers as \emph{quoted answers}, even though the answer is not required to include textual quotation marks around quoted spans. 

We note that some questions might not be adequately answered with this format. Moreover, the feasibility of composing a sufficient answer also depends on the given passages. We argue (and demonstrate), though, that many questions can be answered in a semi-extractive format. This format provides several key advantages described below.

\parhead{Verifiability.}
It is generally easier to verify answers that directly quotes factual statements compared to verifying answers that paraphrase information, even when sentences are attributed to supporting sources. In the latter case, the reader would need to decompose the generated text into statements, review the sources, locate relevant information pieces, and compare the generated statements against them. In contrast, when factual statements are extracted verbatim, verifying the generated text broadly simplifies to ensuring that the statements weren't taken out of context. 
The benefits of quoted spans are even more pronounced in sentences that aggregate information from multiple sources. Rather than citing many sources, leading to combinatorial growth, attributing at a span-level drastically simplifies the verification task.
In \S\ref{sec:results_rating}, we empirically show that this hypothesis holds with human readers.

\parhead{Automatic evaluation.} Another advantage of constraining the output space to include extracted spans is enabling the use of well-defined string-based evaluation metrics for examining both precision and recall of generative models against expert-written references. 
Essentially, this formulation bypasses the ongoing challenge of evaluating free-text answers. Current solutions rely on model-based factual consistency measures~\citep{bohnet2023attributed,honovich-etal-2022-true-evaluating,rashkin2022measuring}.
However, it is hard to establish standardized evaluation protocols with such measures. For example, while larger backbone evaluation models often perform better, sometimes smaller models are used due to  cost and speed considerations. Moreover, evaluating long generated outputs with multiple references exposes even further complexities, as it might require decontextualization~\citep{choi-etal-2021-decontextualization} and aggregating multiple noisy classifications~\citep{laban-etal-2022-summac}. Finally, black-box model-based measures are hard to interpret at an instance-wise level.
In contrast, we introduce text-matching metrics that support multiple sources and are simpler, faster, and easily intepretable.

\parhead{Precision of generated text and attributions.}
Finally, beyond simplifying the verification and evaluation process, using extracted spans from reliable sources for generating factual claims could help prevent generation mistakes and model hallucinations, both in the text itself and in the text-attribution pairing. By enforcing the generative model to explicitly mark the boundaries and the source of each extracted span, the attribution to the correct source is provided by design.

\subsection{Evaluation metrics}\label{sec:eval_metrics}
As discussed above, one of the advantages of SEMQA is in allowing pure string-based evaluation that is interpretable and fast to compute, and does not require additional annotations such as disambiguating questions. 
We formulate the following measures to evaluate answers for fluency, attribution preciseness, and comprehensiveness.

\parhead{Fluency.} Following the ASQA evaluation setting~\citep{stelmakh-etal-2022-asqa}, we use the ROUGE-L score~\citep{lin-2004-rouge} to compare generated answers with reference answers, after stripping any attribution marks. We take the maximum score across all human-written references.

\parhead{Preciseness.} To evaluate the quality of the extracted spans in the generated answer, we compute the normalized token-F1 score~\citep{rajpurkar-etal-2016-squad} for each source separately, and average across sources, taking the maximum score per source across reference quoted answers:

% \begin{small}
\begin{equation*}
    \resizebox{\hsize}{!}{$\operatorname{Sem-F1} (a_i, \mathcal{A}_i) \coloneqq \frac{1}{K} \sum_{k=1}^{K} \max_{\hat{a}\in \mathcal{A}_i}\big(\operatorname{F1}(\psi_k(\hat{a}), \psi_k(a_i))\big)$}
\end{equation*}
% \end{small}
\noindent where $\mathcal{A}_i$ is the set of human-written reference answers for question $q_i$, $K$ is the number of input sources, and $\psi_k(a)$ is a function that keeps only the tokens explicitly marked as extracted from source $k$ in answer $a$. We use the F1 score to measure both the precision of spans (not extracting redundant tokens that don't answer the question), and their recall (extracting all answers and helpful context).

\parhead{Comprehensiveness.} We also measure short-answer recall to capture the aspect coverage of the generated quoted answer:
% While the number of covered answers already impacts the other metrics, we also measure this aspect separately by computing the short-answer recall:

% \begin{small}
\begin{equation*}
    \resizebox{\hsize}{!}{$\operatorname{Sem-Rec} (a_i, \mathcal{S}_i) \coloneqq \frac{1}{K} \sum_{k=1}^{K} \max_{\hat{s}\in \mathcal{S}_{i,k}}\left(\operatorname{Rec}(\hat{s}, \psi_k(a_i)) \right)$}
\end{equation*}
% \end{small}
\noindent where $\mathcal{S}_i$ is the set of short answers,\footnote{We use the provided short answers from the originating PAQ and AmbigQA examples.} by source, covered in each of the human-written references, and $\operatorname{Rec}(\cdot,\cdot)$ is the token-level recall. A perfect score would mean that all short answers appearing in at least one of the references are also in the generated answer, and attributed to the right source. 

\parhead{Combined SEMQA score.} The Fluency score ignores any attribution marks, while Preciseness only measures extraction (i.e., attribution) quality, ignoring the free text portions of the answer. Therefore, we compute their geometric mean, following ASQA~\citep{stelmakh-etal-2022-asqa}, to obtain a single score that reflects the overall answer quality:

% \begin{small}
\begin{equation*}
    \mathrm{SEMQA} \coloneqq (\operatorname{Sem-F1} \cdot \operatorname{ROUGE-L})^{0.5}
\end{equation*}
% \end{small}

\section{The QuoteSum Dataset}\label{sec:dataset}
To study the SEMQA task, we create a new multi-answer QA dataset with $(q, \mathcal{P}, a)$ triplets, where $a$ is a human-written semi-extractive answer. We call this dataset QuoteSum, as it includes quoted answers to the input questions, based on the provided sources. We use square brackets and the index of the respective source to explicitly mark the extracted spans. For example, the first sentence of the answer in Figure~\ref{fig:intro} would be: \mytokens{The song "[ 1 I'll Be Home for Christmas ]" was originally released by [ 1 Bing Crosby ] in [ 1 1943 ]} to mark the spans extracted from source 1. Throughout this paper, we replace the textual marks with colored highlights for presentation.\looseness=-1

Next, in \S\ref{sec:question_colletion} we describe our process for collecting $(q, \mathcal{P})$ pairs based on the PAQ~\citep{lewis-etal-2021-paq} and NQ~\citep{kwiatkowski-etal-2019-natural} datasets. Then, in \S\ref{sec:quotesum} we detail our crowd-sourcing task for writing quoted answers. 
% Lastly, we provide overall dataset statistics in \S\ref{sec:dataset_stats} and define evaluation metrics in \S\ref{sec:eval_metrics}.

\subsection{Collecting multi-answer questions}\label{sec:question_colletion}

\begin{table}[t]
    \centering
    \resizebox{\linewidth}{!}{%\resizebox{\linewidth}{!}{%
\begin{tabular}{l|c}
\toprule
PAQ & Why was the porsche 911 rs built?\ \ \ \qquad \qquad \\
\cline{1-1}
% \multicolumn{2}{c}{ Why do cancer cells develop?}\\ 
\multicolumn{2}{c}{ Where did bob dylan tour in 1978?}\\
\multicolumn{2}{c}{  Which cities in india have police commissionerates?}\\
\midrule
NQ & Who is the actress that portrays wonder woman?\ \ \ \qquad\\
\cline{1-1}
\multicolumn{2}{c}{\ \qquad When did rocky horror picture show come out?}\\
\multicolumn{2}{c}{Who was british pm and viceroy during quit india movement?}\\
\bottomrule
\end{tabular}
}%
\vspace{-5pt}
    \caption{Example questions from \thedataset dataset, requiring aggregating information from multiple sources.}
    \label{tab:question_examples}
    \vspace{-5pt}
\end{table}

Since we focus on multi-answer questions, we first need to collect the questions and candidate answers that will be provided to the writers for producing the semi-extractive answers. We obtain both machine-generated questions from the PAQ dataset~\citep{lewis-etal-2021-paq} and human-written questions from the AmbigQA~\citep{min-etal-2020-ambigqa} subset of NQ~\citep{kwiatkowski-etal-2019-natural}. Table~\ref{tab:question_examples} presents example selected questions.

\parhead{Probably-Asked Questions (PAQ).}
Our first set of questions is based on the PAQ dataset~\citep{lewis-etal-2021-paq}, a collection of 65 million probably-asked questions that were automatically generated from Wikipedia passages. For each passage, possible answer spans are extracted with a BERT~\citep{devlin-etal-2019-bert} model. Then, a question is generated with a BART-base~\citep{lewis-etal-2020-bart} question generation model trained on NQ~\citep{kwiatkowski-etal-2019-natural}, TriviaQA~\citep{joshi-etal-2017-triviaqa}, and SQuAD~\citep{rajpurkar-etal-2016-squad}.

We start with the large PAQ collection and look for multi-answer questions, meaning instances where the same question was independently generated from two different target answers and passages. Therefore, we merge all $(q, p, s)$ triplets of question, passage, and short answer by identical questions to obtain a mapping $q \rightarrow [(p_1, s_1), \ldots, (p_k, s_k)]$. 

To avoid including multiple instances of the same answer, we apply a series of filters.\footnote{The initial large size of this dataset allows us to filter aggressively as we can afford false positives.} For each question $q$, we sort the $(p_i, s_i)$ by $\mathbb{P}(s_i \mid q, p_i)$ according to a T5-XXL~\citep{t5paper} QA model trained on SQuAD, and keep only answers with score of at least $0.5$. Then, we run over $(p_i, s_i)_{i=1}^{k}$ and filter out any instances where either (1) the Wikipedia page of $p_i$ was already included; (2) the length of $s_i$ is less than 4 words; (3) $s_i \in p_j$ for $j<i$; or (4) $\phi(s_i, s_j)=1$ for $j<i$. 

The function $\phi(x,y)$ is a binary answer similarity function that returns $1$ if at least one of the following conditions are true: (1) the Levenstain distance $\mathrm{Lev}(x,y) \le 10$; (2) the word-based intersection-over-union $\mathrm{IoU}(x,y) > 0.75$; or (3) the score of a semantic answer-similarity classifier $\mathrm{BEM}(x,y)>0.5$~\citep{bulian-etal-2022-tomayto}. The first term is useful for identifying different spellings of the same name (e.g., Céline and Celine). The second term helps for finding lists that are ordered differently. The last semantic BEM-based term catches other more subtle duplicates.

We also find that some slightly different questions are similar in meaning. To automatically merge similar questions, we vectorize them with TF-IDF and merge questions with cosine similarity greater than $0.9$. When merging instances, we repeat the same answer filtering process above, and keep the question obtaining higher average score against the answers, by the same T5 QA model.

This process resulted in about $170K$ questions, each with at least two answer passages. At first, the questions were highly unbalanced in type (e.g., many \textit{what} and \textit{who} questions) and in the number of answers. Therefore, we partition the data by questions starting with \textit{what, who, where, when, how, which, why}, questions including \textit{stand for} or ``other''. We also partition by the number of answers. Then, we perform balanced random sampling when selecting the examples for \thedataset.

\parhead{Natural Questions (NQ).} We also collect human-written questions based on the NQ dataset~\citep{kwiatkowski-etal-2019-natural}. Since we are interested in multi-answer questions, we use the subset of questions identified by \citet{min-etal-2020-ambigqa} to be multi-answer, together with their annotated short answers $s_i$ to disambiguating questions. We also collect the Wikipedia passages $p_i$  that contain $s_i$. We notice that some disambiguating questions $q'_i$ could be slightly nuanced. To collect passages that directly answer the original questions $q$, we filter out $(p_i, s_i)$ pairs for question $q$ where $\mathbb{P}(s_i \mid q, p_i)<0.5$ according to a T5-XXL SQuAD-trained model.

Since some short answers to the same question are coming from the same passage, we remove passages that have word intersection ratio greater than $0.4$ with another passage. Finally, we keep only questions with at least two different answers.

Similar to the PAQ set, we see an imbalance in the question types (with many \textit{who} and \textit{when} questions). Again, we balance the sampling across question types and number of answers to diversify the examples to include in \thedataset.

\subsection{Writing quoted answers}\label{sec:quotesum}
We now detail our annotation task for creating human-written semi-extractive answers $a$ for our gathered questions and passages $(q, \mathcal{P})$. We build a web interface for writers that presents a question and a list of passages (see Figure~\ref{fig:writing_plugin}). Writers can select spans from input passages and copy them to the answer text box. These copied spans are colored to match the respective input source to help the writer keep track of the covered sources. Writers can also use free text to connect the extracted spans and complete the answer. The submitted answers include our special marks for signaling which spans were copied, and their originating passage.

Writers are asked to create a concise summary that answers the question while copying informative spans from the sources whenever possible, and using free text with a diverse set of connectors. They are told to cover as much information as possible, as long as it directly relates to the given question. See Appendix~\ref{app:write_task} for more details.

We only provide the writers with the questions and matching passages, without revealing any short answer spans. We choose so to avoid biasing the writers to follow any noisy answers that might have been included through the scraping process. Therefore, writers can choose not to use all input passages if they find some to be irrelevant. We view this selection process as part of the task, and keep all input passages in our dataset, realistically simulating a slightly noisy retrieval output provided to the answer generation model.

\subsection{Dataset statistics}\label{sec:dataset_stats}
% In total,
% \thedataset includes 4,009 semi-extractive answers to 1,376 unique questions (984 from PAQ and 392 from NQ). As mentioned in \S\ref{sec:question_colletion} we balance the data via question sampling to increase diversity. Question type and number of input passages statistics are provided in Figure~\ref{fig:stats}. 
% We split the data to train, validation, and test sets with ratios 60\%, 7\%, 33\%, respectively.
% We ensure that the set of Wikipedia pages that the test set answers are based on are disjoint from the train and validation sets.

\begin{figure}[t]
     \centering
     \begin{subfigure}[b]{0.4\textwidth}
         \centering
    \includegraphics[width=\textwidth]{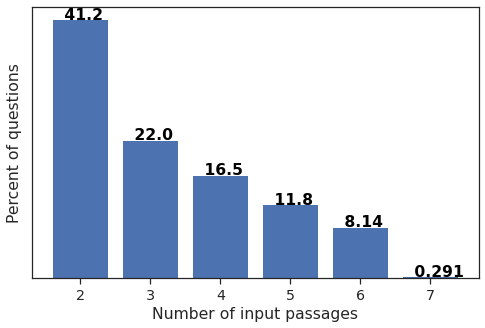}
        %  \caption{$y=x$}
        %  \label{fig:y equals x}
     \end{subfigure}
     \begin{subfigure}[b]{0.4\textwidth}
         \centering
    \includegraphics[width=\textwidth]{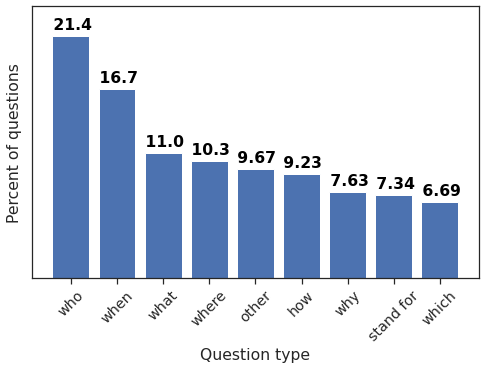}
        %  \caption{$y=5/x$}
        %  \label{fig:five over x}
     \end{subfigure}
     \vspace{-2pt}
    \caption{Diversity of questions in \thedataset.}
    \label{fig:stats}
    \vspace{-6pt}
\end{figure}

\thedataset includes 4,009 semi-extractive answers to 1,376 unique questions (984 from PAQ and 392 from NQ). As reported in Figure~\ref{fig:stats} and illustrated in Table~\ref{tab:question_examples}, \thedataset includes a diverse set of multi-answer questions. See App.~\ref{app:dataset_stats} for more details.

% \begin{figure}[t]
%     \centering
%     \includegraphics[width=0.4\textwidth]{figures/quotesum_n_stats2.png}
%     \caption{Input passages per question in \thedataset.}
%     \label{fig:stats_n}
% \end{figure}
% \vspace{-4pt}

% \begin{figure}[t]
%     \centering
%     \includegraphics[width=0.4\textwidth]{figures/quotesum_q_stats2.png}
%     \caption{Question type statistics in \thedataset.}
%     \label{fig:stats_q}
% \end{figure}
% \vspace{-4pt}

\section{Experimental Evaluation}\label{sec:exp_eval}
We study the performance of both fine-tuned LMs, and in-context few-shot LLMs, as described below. We follow the evaluation metrics described in \S\ref{sec:eval_metrics}, and also run a user study as detailed in \S\ref{sec:results_rating}.

\parhead{Fine-tuned models.}
We fine-tune different sizes of T5 models~\cite{t5paper} on the \thedataset training set. We also fine-tune FLAN instruction-tuned T5 models~\citep{flan_paper}. See Appendix~\ref{app:imp_details} for more details.

\parhead{In-context learning.}
We evaluate variants of few-shot PaLM2~\citep{palm2}. We construct a dynamic prompt that collects examples for each test question $q_i$ by retrieving the most similar questions from the \thedataset training set according to Sentence-T5~\citep{ni-etal-2022-sentence} embedding cosine similarity (QSum). We also experiment with the ALCE~\citep{Gao2023EnablingLL} prompt that uses fully-abstractive answers with in-line citations (e.g., [1] to cite source 1), and with Qsum-S prompts that convert QSum examples to sentence-level citations. See Appendix~\ref{app:imp_details} for details and example prompts.

\subsection{Results and analysis}\label{sec:main_results}
Table~\ref{tab:results_tuned} summarizes the results of the fine-tuned models and the in-context learners.
Fine-tuned models generally perform better at generating quoted answers as they increase in size, both in terms of short-answer coverage (Sem-Rec), and in terms of fluency (Rouge-L) and preciseness of the extracted spans (Sem-F1). Also, the Flan-T5 model family is significantly better than the original T5, possibly due to the set of instruction-following tasks they were trained on that might have enabled better adaptation to new output formats. For example, Flan-T5 base is matching the scores of T5 XL that has about $\times 13$ more parameters.

The few-shot learners obtain relatively high performance with only a few examples, demonstrating their sample efficiency. However, they still lack behind the fine-tuned models, suggesting the value of our quoted summaries for training models to generate consolidated quoted answers. 

\parhead{Interpreting the string-based metrics.} Our metrics can help understand the strengths and weaknesses of different models. For example, 5-shot models obtain reasonably high Sem-Rec score, with PaLM2 Unicron achieving as high as 88.87, meaning that answers successfully cover multiple aspects of the question. The lower Sem-F1 scores show that the model can be more precise in extracting only relevant spans for answering the question. 

Also, comparing the few-shot models to the T5 large model, we see that the Rouge-L scores are comparable while the Sem-F1 score of the T5 model is much higher. This suggests that the few-shot model matched the fluency of the T5 model, but didn't mark all extracted spans as quotes.

This is demonstrated in Table~\ref{tab:example_answer} with an example answer by PaLM2 Bison 4-shot. The  quoted answer comprehensively extracts all short-answers from the given passages while also helpfully contextualizing each option. While the extracted spans provide useful attributions to the supporting passages, the preciseness of this output can be improved. For example, ``Susan Egan'' can be dropped, and the year of each show that is given in free-text should instead be quoted from the respective passage, like in the human-written reference.

\parhead{String-based vs.\ model-based metrics.}
For comparison, we also report the model-based evaluation metrics from ASQA~\citep{stelmakh-etal-2022-asqa}. The Disambig-F1 metric requires running an extractive RoBERTa large~\citep{liu2019roberta} QA model on a list of disambiguating questions and the generated long answer, and comparing each predicted answer span with the gold short-answer. Since we don't have disambiguating questions for PAQ examples, we only evaluate the portion of our test set originating from AmbigQA. As Table~\ref{tab:results_disambig} shows, our Sem-F1 text-based measure ranks the examined models the same as Disambig-F1, without relying on any QA models or disambiguating questions.

\begin{table}[t]
    \centering
    \small
\resizebox{\linewidth}{!}{%
\begin{tabular}{ldddd}
\toprule
Model &  \hd{Rouge-L} &  \hd{Sem-F1} &  \hd{Sem-Rec} &  \hd{SEMQA} \\
\midrule
T5 small     &      54.29 &         57.66 &   68.30 &     55.95 \\
T5 base      &      60.39 &         67.69 &   78.04 &     63.94 \\
T5 large     &      63.90 &         71.55 &   83.68 &     67.61 \\
T5 XL        &      65.36 &         73.05 &   84.88 &     69.10 \\
T5 XXL       &      70.49 &         80.60 &   91.25 &     75.38 \\
\midrule
Flan-T5 small&      58.33 &         63.05 &   78.24 &     60.65 \\
Flan-T5 base &      65.08 &         73.55 &   86.32 &     69.19 \\
Flan-T5 large&      71.59 &         81.62 &   91.79 &     76.44 \\
Flan-T5 XL   &      72.89 &         83.82 &   94.38 &     78.17 \\
Flan-T5 XXL  &      73.36 &         84.20 &   93.68 &     \textbf{78}.\textbf{59} \\
\midrule
PaLM2 Bison 1-shot             &      60.04 &         50.11 &   52.62 &     54.85  \\
PaLM2 Bison 4-shot             &      62.71 &         62.11 &   77.32 &     62.41 \\
PaLM2 Bison 5-shot             &      63.45 &         65.40 &   80.09 &     64.42 \\
PaLM2 Unicorn 5-s.             &      64.13 &         68.28 &   88.87 &     66.18 \\
\bottomrule
\end{tabular}

}%

    \caption{Test results of fine-tuned and few-shot models.}
    \label{tab:results_tuned}
\end{table}

\begin{table}[]
    \centering
    \resizebox{\linewidth}{!}{%\resizebox{\linewidth}{!}{%
\begin{tabular}{cllddd|dd}
\toprule
Ver. & $n$ & Prompt &  \hd{Rouge} &  \hd{Sem-F1} &  \hd{SEMQA} &  \hd{Dis.-F1} &    \hd{DR} \\
\midrule
Bis. & 4 & QSum            &      64.46 &   58.04 &     61.17 &        45.32 & 54.05 \\
Bis. & 5 & QSum            &      61.75 &   59.15 &     60.44 &        46.06 & 53.33 \\
Uni. & 5 & QSum                    &      64.23 &   63.76 &     64.00 &        \textbf{49}.\textbf{05} & \textbf{56}.\textbf{13} \\
\midrule
Bis. & 4 & QSum-S  &      60.17 & -   &    -  &        39.03 & 48.46 \\
Bis. & 4 & ALCE &      54.73 &   - &     - &        34.05 & 43.17 \\
\bottomrule
\end{tabular}
}%

    \caption{PaLM2 $n$-shot results on the AmbigQA~\citep{min-etal-2020-ambigqa} subset of \thedataset test set using our \thedataset prompt (QSum), ALCE prompt~\citep{Gao2023EnablingLL}, and \thedataset prompt converted to sentence-level citations (QSum-S). We also report the ASQA RoBERTa-based Disambig-F1 score and the combined DR score~\citep{stelmakh-etal-2022-asqa}.}
    \label{tab:results_disambig}
\end{table}

\begin{table*}[t]
\centering
\begin{small}
% \begin{minipage}{\linewidth}
        \resizebox{\linewidth}{!}{
% \begin{tabular}{lp{0.15\linewidth}p{0.4\linewidth}p{0.08\linewidth}p{0.12\linewidth}r}
\begin{tabular}{p{0.06\linewidth}|p{1.00\linewidth}}
\toprule
Question & \textit{Who plays as beast in beauty and the beast?} \\
\midrule
% \multicolumn{2}{l}{\emph{...:}} \\
Sources & \tcbox[colback = hl_1]{[1]} Beauty and the Beast (\tcbox[colback = hl_1]{1987 TV series}): Beauty and the Beast is an American fantasy-drama television series which first aired on CBS from September 25, 1987 to August 4, 1990. Creator Ron Koslow's updated version of the fairy tale has a double focus: the relationship between Vincent (\tcbox[colback = hl_1]{Ron Perlman}), a mythic, noble man-beast, and Catherine (Linda Hamilton), a [...] \\
& \tcbox[colback = hl_2]{[2]} Beauty and the Beast (\tcbox[colback=hl_2]{musical}): After completing tryouts in Houston, "Beauty and the Beast" \tcbox[colback=hl_2]{premiered on Broadway on April 18, 1994}, starring Susan Egan and \tcbox[colback=hl_2]{Terrence Mann} as the eponymous Belle and Beast, respectively. The \tcbox[colback=hl_2]{musical} opened to mixed reviews from theatre critics, but was a massive commercial success [...] \\
& \tcbox[colback = hl_3]{[3]} Beauty and the Beast (1991 film): "Beauty and the Beast" focuses on the relationship between the Beast (voice of Robby Benson), a prince who is magically transformed into a monster and his servants into household objects as punishment for his arrogance, and Belle (voice of Paige O'Hara), a young woman whom he imprisons in his castle to become a [...]\\
& \tcbox[colback = hl_4]{[4]} Beauty \& the Beast (\tcbox[colback=hl_4]{2012 TV series}): Kristin Kreuk and \tcbox[colback=hl_4]{Jay Ryan} star in the title roles alongside Austin Basis, Nina Lisandrello, Nicole Gale Anderson, Sendhil Ramamurthy, Max Brown, Brian J. White, Amber Skye Noyes, and [...] \\
& \tcbox[colback = hl_5]{[5]} Beast (Beauty and the Beast): In all animated film appearances, the Beast is voiced by American actor  \tcbox[colback=hl_5]{Robby Benson}.  \tcbox[colback=hl_5]{The 1991 animated film} was adapted into a Broadway musical in 1994, with the role being originated by American actor Terrence Mann. \tcbox[colback=hl_5]{Dan Stevens} portrays a live-action version of the character in \tcbox[colback=hl_5]{the 2017 live-action adaptation} of the original 1991 film. \\
& \tcbox[colback = hl_6]{[6]} Beauty and the Beast (\tcbox[colback=hl_6]{2014 film}): On the way home through the forest, the merchant loses his way, his horse slips and is injured, and they are attacked by wolves. He laments that he has "not even a weapon to finish off" the poor horse. The merchant stumbles upon the magical domain of the Beast (\tcbox[colback=hl_6]{Vincent Cassel}).\\
& \tcbox[colback = hl_7]{[7]} Beauty and the Beast (\tcbox[colback=hl_7]{1946 film}): Awakened by a loud roar, he wanders the castle's grounds. Remembering that Belle asked for a rose, he plucks a rose from a tree which makes the Beast (\tcbox[colback=hl_7]{Jean Marais}) appear. The Beast threatens to kill him for theft but suggests that one of his daughters can take his place.\\
\midrule
SEMQA (human-written) & The actors who play the Beast in the various versions of Beauty and the Beast include \tcbox[colback=hl_5]{Dan Stevens} in \tcbox[colback=hl_5]{the 2017 live-action adaptation} ; \tcbox[colback=hl_6]{Vincent Cassel} in the \tcbox[colback=hl_6]{2014 film} ; \tcbox[colback=hl_4]{Jay Ryan} in the \tcbox[colback=hl_4]{2012 TV series} ; \tcbox[colback=hl_2]{Terrence Mann} in the \tcbox[colback=hl_2]{musical} that \tcbox[colback=hl_2]{premiered on Broadway on April 18, 1994} ; \tcbox[colback=hl_5]{Robby Benson} in \tcbox[colback=hl_5]{The 1991 animated film} ; \tcbox[colback=hl_1]{Ron Perlman} in the \tcbox[colback=hl_1]{1987 TV series} and \tcbox[colback=hl_7]{Jean Marais} in the \tcbox[colback=hl_7]{1946 film}.\\
\midrule
SEMQA (model) & \tcbox[colback=hl_1]{Ron Perlman} plays as \tcbox[colback = hl_1]{Beast} in \tcbox[colback = hl_1]{Beauty and the Beast} (1987 TV series). \tcbox[colback=hl_2]{Susan Egan} and \tcbox[colback=hl_2]{Terrence Mann} play as \tcbox[colback=hl_2]{Belle} and \tcbox[colback=hl_2]{Beast} respectively in \tcbox[colback=hl_2]{Beauty and the Beast} (musical). \tcbox[colback=hl_3]{Robby Benson} is the \tcbox[colback=hl_3]{voice} of \tcbox[colback=hl_3]{Beast} in \tcbox[colback=hl_3]{Beauty and the Beast} (1991 film). \tcbox[colback=hl_4]{Kristin Kreuk} and \tcbox[colback=hl_4]{Jay Ryan} star in the title roles in \tcbox[colback=hl_4]{Beauty \& the Beast} (2012 TV series). \tcbox[colback=hl_5]{Dan Stevens portrays} a live-action version of the character in the 2017 live-action adaptation of the original 1991 film. \tcbox[colback=hl_6]{Vincent Cassel} plays as \tcbox[colback=hl_6]{Beast} in \tcbox[colback=hl_6]{Beauty and the Beast} (2014 film). \tcbox[colback=hl_7]{Jean Marais} plays as \tcbox[colback=hl_7]{Beast} in \tcbox[colback=hl_7]{Beauty and the Beast} (19 [...]\\
\midrule

ALCE (model) & The Beast is played by Ron Perlman in the 1987 TV series [1], Terrence Mann in the 1994 musical [2], Robby Benson in the 1991 film [3], Jay Ryan in the 2012 TV series [4], Vincent Cassel in the 2014 film [6], and Jean Marais in the 1946 film [7].\\
\bottomrule
\end{tabular}
% \end{minipage}
}%
\vspace{-4pt}
\caption{Human-written example answer from \thedataset, and generated answers of 4-shot PaLM2 Bison with our SEMQA prompt, and with the ALCE prompt. Colored spans show the spans that the writers or the model marked as explicit extractions from the respective source, by color. The extracted spans for the human-written answer are also highlighted in the input sources.}\label{tab:example_answer}
\end{small}
\end{table*}

\parhead{Comparing to answers with citations.}\label{sec:results_rating}
One alternative approach for attributing answers is to generate the answer text with citations. We evaluate SEMQA against that approach using both automatic metrics and human ratings. To this end, we compare against the recently proposed few-shot ALCE method~\citep{Gao2023EnablingLL} that, similar to us, prompts an LLM with the question and a set of pre-retrieved passages. We use the ALCE ASQA light instruction prompt\footnote{\url{https://github.com/princeton-nlp/ALCE}} that includes 4 examples of fully-abstractive answers with in-line citations to input sources.

\begin{figure}[t]
    \centering
    \includegraphics[width=0.48\textwidth]{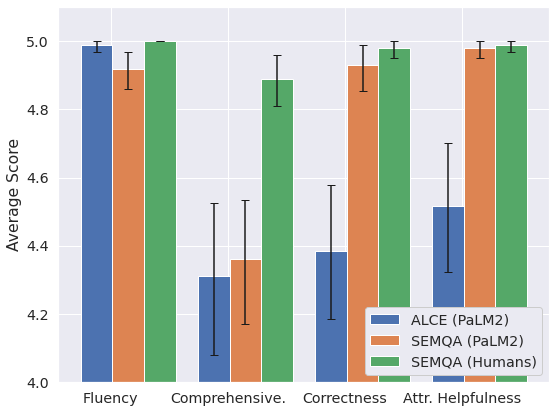}
    \caption{Average human ratings (1-5 scale) for answers of PaLM2 Bison 4-shot prompted with either SEMQA or ALCE examples, and for human-written SEMQA answers from QuoteSum. Error bars show 95\% confidence intervals obtained with bootstrapping.}
    \label{fig:rating_semqa_alce}
\end{figure}
% \vspace{-6pt}

\begin{figure}[t]
    \centering
    \includegraphics[width=0.4\textwidth, height=100pt]{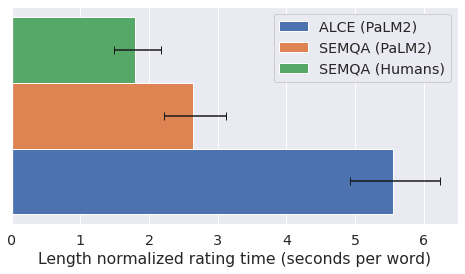}
    \caption{Manually evaluating and verifying generated SEMQA answers is twice faster than evaluating fully abstractive answers. Error bars show the 95\% confidence intervals obtained with bootstrapping.}
    \label{fig:rating_time_semqa_alce}
    % \vspace{-4pt}
\end{figure}

Table~\ref{tab:results_disambig} reports the ASQA metrics~\citep{stelmakh-etal-2022-asqa} for PaLM2 with our \thedataset prompt and the ALCE prompt. Since in ALCE spans are not explicitly quoted, even if copied from the input passages, they cannot be scored with our semi-extractive metrics. The \thedataset prompt (QSum) leads to higher scores, both in ROUGE-L and in Disambig-F1, suggesting that beyond better interpretability, our semi-extractive approach also provides more precise answers. To more directly compare the two approaches, we automatically (using a regex function) convert our \thedataset examples to remove span-level attributions and instead add respective sentence level citations (QSum-S). Then, instead of the ALCE 4 examples, we dynamically retrieve examples from the training set by question similarity (like in QSum). While improving over the ALCE prompt, the converted format (QSum-S) still scores lower than the semi-extractive format (QSum), indicating the benefits of explicitly extracting spans.

We also run a user study where human raters (a different pool than the one creating \thedataset) were asked to score generated answers for fluency, comprehensiveness, correctness, and attribution helpfulness, using a 5-level Likert scale (See Appendix~\ref{app:rating_task} for details).\footnote{We choose scoring over side-by-side evaluation due to the clear guidelines and factual nature~\citep{bansal2024peering}.} We compare the generated answers of PaLM2 Bison 4-shot prompted with either SEMQA or ALCE examples, and SEMQA human-written answers from \thedataset. As Figure~\ref{fig:rating_semqa_alce} shows, ratings were overall high for both models, yet still below human performance. The SEMQA model scored slightly lower than ALCE on fluency, indicating the challenge in ``glueing'' the extracted spans in a coherent way. However, the human-written references obtained virtually perfect fluency score, demonstrating the potential of semi-extractive fluent answers.

The SEMQA model performed better on comprehensiveness, and significantly better on correctness, and attribution helpfulness. This supports our hypothesis that semi-extractive generation reduces hallucinations, and that span-level attributions help the reader verify the generated text.

\parhead{Human verification with SEMQA is easier.}
We also measure the rating time for each example. Since SEMQA answers were generally longer than ALCE answers, we normalize by the number of words in each answer (excluding any attribution marks) to compute the average seconds per word. As shown in Figure~\ref{fig:rating_time_semqa_alce}, rating generated SEMQA answers was more than twice faster than rating ALCE abstractive answers. This suggests that in addition to being higher-quality answers (according to ratings), SEMQA answers with clearly marked quotes are also easier to evaluate and verify by readers.

% Please add the following required packages to your document preamble:
% \usepackage{graphicx}
\begin{table}[]
\small
\resizebox{\columnwidth}{!}{%
\begin{tabular}{l|dddd}

\toprule
Baseline       &    \hd{Rouge} &  \hd{Sem-F1}  & \hd{Sem-Rec} &  \hd{SEMQA} \\
\midrule
Lead 1        & 29.58       & 30.01      & 39.99       & 29.79    \\
Lead 2        & 34.32       & 32.74      & 60.36       & 33.52    \\
Lead 3        & 33.69       & 34.96      & 77.44       & 34.32    \\
Lead 4        & 31.52       & 33.45      & 88.55       & 32.47    \\
Lead 5        & 29.87       & 31.96      & 93.55       & 30.90    \\
\midrule
Tail 1         & 23.79       & 18.47      & 41.22       & 20.96    \\
Tail 2         & 28.26       & 27.14      & 61.10       & 27.69    \\
Tail 3         & 28.63       & 29.35      & 79.97       & 28.99    \\
Tail 4         & 28.61       & 30.06      & 88.80       & 29.32    \\
Tail 5         & 28.39       & 30.19      & 94.56       & 29.27    \\    
\bottomrule
\end{tabular}%
}
\caption{Trivial baselines that simply extract the first or last $k$ sentences from each source perform poorly.} \label{tab:naive_baselines}
\end{table}

\parhead{Solving SEMQA is non-trivial.}
The results of our experiments and user-study show that both finetuned and in-context LLMs can achieve noteworthy performance on SEMQA, thanks to our \thedataset dataset and task formulation. However, the results also show that current models still lack behind human performance, indicating the challenges of this task. To further validate the complexity of this task and dataset, we also verify that trivial solutions cannot solve it or trick our string-based metrics. 

Previous studies have found naïve baselines such as taking the first few sentences of the source document to perform well on certain extractive summarization datasets~\citep{grenander-etal-2019-countering,zhong-etal-2019-closer}. Table~\ref{tab:naive_baselines} reports the \thedataset performance of a baseline that extracts the first or last $k$ sentences from each source and simply concatenates them as the answer. 
As expected, these baselines perform poorly (compared to all models; see Table~\ref{tab:results_tuned}). Unsurprisingly the Sem-Rec score, that only measures recall of short answers, increases with more sentences included. However, the Rouge-L, Sem-F1, and the aggregated SEMQA scores are very low as they also measure the precision of the selected spans and the fluency of the answer compared to the references.

\section{Conclusion}\label{sec:conclusion} 
We introduce the task of semi-extractive multi-source QA (SEMQA), and a new comprehensive dataset with human-written SEMQA examples (\thedataset) with text-based evaluation metrics. Our experiments and analysis reveal the benefits of SEMQA for enabling LLMs to generate well-grounded answers that are fast to verify against external sources, and easy to interpret and evaluate against references---avoiding the reliance on model-based measures.
We hope that SEMQA and the released data will facilitate further research on improving LLM capabilities around reliable QA.

We make the \thedataset dataset and implementation of the SEMQA metrics available at: \url{https://github.com/google-research-datasets/QuoteSum}.

\section*{Acknowledgments}
We would like to thank Kai Hui, Dipanjan Das, Mirella Lapata, Shashi Narayan, Kellie Webster, Vinh Q. Tran, Gaurav Singh Tomar, Michael Collins, Mor Geva, Roee Aharoni, Irene Ros, Jing Wang, Jayant Madhavan, John Blitzer and many other Googlers for fruitful discussions that helped shape this research project. We also thank the crowd-sourcing team for their help with the dataset creation and the user study.

\section*{Limitations}
Our study is limited in scope to English questions and answers, and to English Wikipedia articles as supporting sources. While we applied diversified balanced sampling for choosing multi-answer questions out of the larger NQ and PAQ collections, post our filtering and merging functions, the scope of our examined questions is still limited to the ones contained in those collections. Also, as we discuss throughout the paper, we don't aim to suggest that our SEMQA approach is ideal for all types of questions and answers. Rather, we argue that it is suitable for many multi-answer questions, and that it provides many benefits when appropriately followed.

In our dataset and experiments, we provide a set of retrieved sources that we collect following the process described in section~\ref{sec:dataset}. This setup is useful for objective evaluation of models, focusing on the reading comprehension and information consolidation capabilities. Many other notable QA datasets such as SQuAD~\citep{rajpurkar-etal-2016-squad} and Natural Questions~\citep{kwiatkowski-etal-2019-natural} also first focused on such setups, leaving extensions to open-domain systems to later work~\citep{chen-etal-2017-reading}. One limitation of the closed-domain setup is that models might fail properly generalize to inputs from noisy retriever systems. However, as we discuss in the paper, our data collection process intentionally kept some noisy sources in the input which were not cited by the writers in the reference answers. Therefore, to perform well on \thedataset, models should already learn to filter out irrelevant sources. Moreover, it is generally easier to further augment the dataset automatically with unused sources.

We also clarify that while the semi-extractive format might increase alignment with the supporting attributed sources, it is not bullet-proof against model hallucinations such as incorrect coreferences or other out-of-context issues~\citep{zhang2023extractive}. We find semi-extractive answers to be easier to manually verify against the attributed source. Finally, it is important to note that the correctness of the generated answer also depends on the accuracy of the provided sources, which might include incorrect or misleading information. Here, to simplify the setting and keep it focused, we assume that the sources are provided by some possibly noisy retriever. We leave further studies on the retrieval part to future studies.

\section*{Broader Impact}
We publicly release the dataset collected for this study, including the human-written semi-extractive answers. We hope that this will facilitate future research on enabling semi-extractive QA systems, further exploring their tradeoffs, and identifying the ideal question coverage for such models.

% Entries for the entire Anthology, followed by custom entries
\bibliography{anthology,custom}

% \clearpage
\appendix
\counterwithin{figure}{section}
\counterwithin{table}{section}

\section{Details of Answer Writing Task}
\label{app:write_task}

\begin{figure*}[t]
    \centering
    \includegraphics[width=1.0\textwidth]{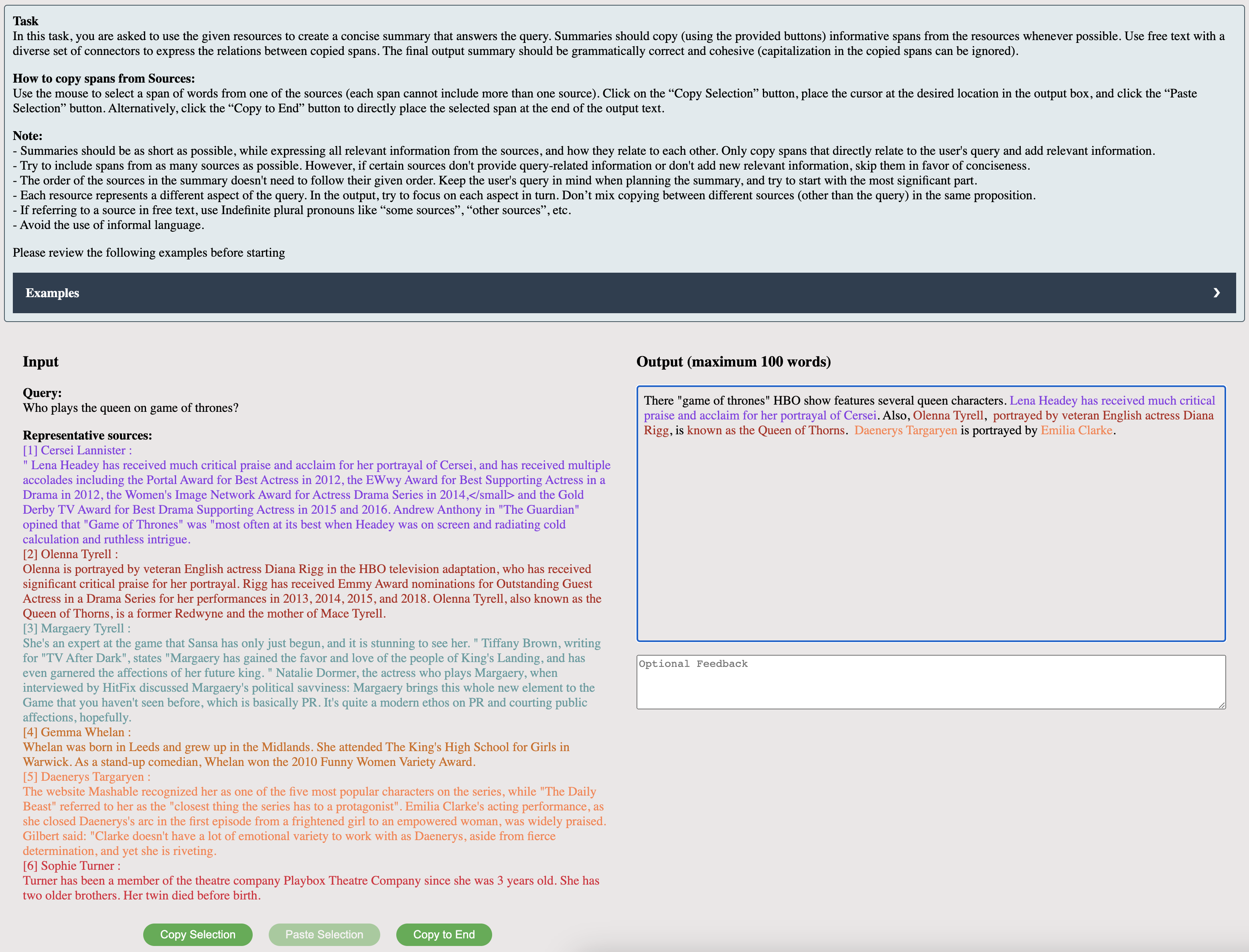}
    \caption{A screenshot of our tool and short instructions for writing semi-extractive answers.}
    \label{fig:writing_plugin}
\end{figure*}

We crowd-sourced native English speakers with Bachelor's degrees in Journalism or related fields to compile the semi-extractive answers as described in \S\ref{sec:quotesum}.
A screenshot of the writing plugin used for this task is shown in Figure~\ref{fig:writing_plugin}. We constricted the answer to be no more than 100 words in order to keep it concise and query-focused. We find this limit to be sufficient for properly answering the examined questions.

Before starting the main writing task, we performed three pilot runs in which writers worked on 10 example questions. After each pilot run, we examined the written answers, provided feedback, refined the instructions, and also provided further reference example answers that the writers could learn from. Writers were also given the option to provide feedback to specific questions, which allowed us to discuss and clarify any confusions regarding the task. 

The main writing task was performed in batches. After each batch, we randomly examine a subset of the answers and provide additional feedback and clarifications when needed. For instance, after the first batch, we identified that one of the writers produces overly-extractive answers that are not fluent. We sent these answers back for rewriting, and clarified the guidelines accordingly.

\section{Details of the User Study}
\label{app:rating_task}

In the generated answer rating task, raters were given a single example each time that includes the question, the sources, and the machine-generated answer. We rendered the outputs to highlight the extracted spans in SEMQA answers with colors that match the respective spans in the input sources (similar to the visualization demonstrated in Figure~\ref{fig:intro}). For consistency across the two formats, we also post-process the SEMQA outputs (with a regular expression function) to add sentence-level citations similar to ALCE. Raters were only told that attributions are provided as citations, and sometimes also as highlighted colored spans that match respective spans in the sources. All examples were shuffled and raters were not told they can originate from different models.

We used a different pool of raters than the one used for the QuoteSum dataset. We asked raters to mark their agreement with the following statements, using a five-level Likert scale:
\begin{enumerate}[leftmargin=*,noitemsep]\vspace{-4pt}
    \item \textbf{Fluency}: The machine-generated answer is fluent and coherent.
    \item \textbf{Comprehensiveness}: The machine-generated answer is comprehensive in covering different aspects of the question.
    \item \textbf{Correctness}: All information in the machine-generated answer is fully supported by the provided respective attributions.
    \item \textbf{Attribution helpfulness}: The attributions in the machine-generated answer are helpful for evaluating its correctness.
\end{enumerate}

We collected ratings for 100 outputs of each model with 3 repetitions by independent raters. Inter-rater agreement was high ($0.75$ Krippendorff's alpha). 

\begin{figure*}[t]
    \centering
    \includegraphics[width=1.0\textwidth]{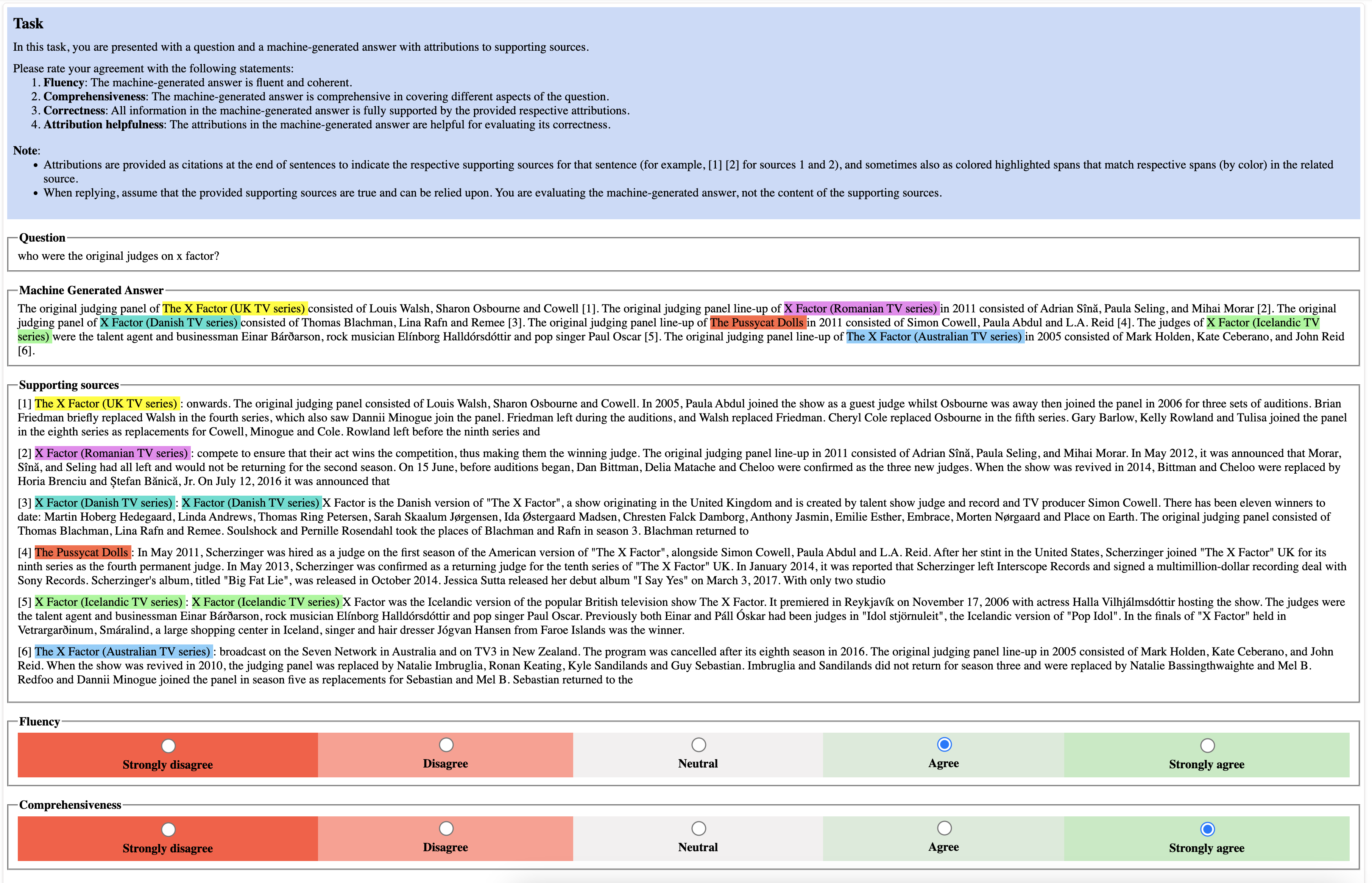}
    \caption{A screenshot of our tool for rating the quality of generated answers against the supporting sources.}
    \label{fig:rating_plugin}
\end{figure*}

A screenshot of the rating plugin used for human evaluation of machine-generated answers is shown in Figure~\ref{fig:rating_plugin}.

\section{Additional Dataset statistics}\label{app:dataset_stats}

% \begin{figure}[t]
%      \centering
%      \begin{subfigure}[b]{0.4\textwidth}
%          \centering
%     \includegraphics[width=\textwidth]{figures/quotesum_n_stats2.png}
%         %  \caption{$y=x$}
%         %  \label{fig:y equals x}
%      \end{subfigure}
%     %  \begin{subfigure}[b]{0.4\textwidth}
%     %      \centering
%     % \includegraphics[width=\textwidth]{figures/quotesum_q_stats2.png}
%     %     %  \caption{$y=5/x$}
%     %     %  \label{fig:five over x}
%     %  \end{subfigure}
%      \vspace{-2pt}
%     \caption{Distribution of \thedataset examples by number of input passages.}
%     \label{fig:stats_app}
%     \vspace{-6pt}
% \end{figure}

As mentioned in Section~\ref{sec:dataset_stats}, \thedataset includes a diverse set of multi-answer questions with different question types. As Figure~\ref{fig:stats} shows, the examples also vary in the number of provided input passages. The maximum number of input passages in the dataset is seven, and most examples have two or three input passages.

We split the data to train, validation, and test sets with ratios 60\%, 7\%, 33\%, respectively.\footnote{We use a training split to enable fair comparison between finetuned and in-context learning (ICL) models. However, studies that only focus on ICL models might choose to use most or all of the collected dataset for evaluation.}
We ensure that the set of Wikipedia pages that the test set answers are based on are disjoint from the train and validation sets.

\section{Implementation Details}
\label{app:imp_details}

To fine-tune the T5 models~\citep{t5paper} we use the T5X framework~\citep{roberts2022t5x}. We use the T51.1 and Flan-T5 checkpoints and train the models on TPUv4 chips, other than the ``large'' or smaller Flan-T5 models that used TPUv3 chips. We trained the models for up to 25k steps with a batch size of 32 and learning rate of 1e-3 with the Adafactor optimizer~\citep{shazeer2018adafactor}. For reference, training Flan-T5 base on 4 TPUv3 took 1.5 hours. For each model, we picked the best checkpoint according to the ROUGE-L score on the validation set. We did not perform any hyper-parameter tuning, and only retrain the T5-XL once with a new random seed, since the first training run of that model led to low training and validation scores. 

Our few-shot evaluations are querying variants of the PaLM2 model~\citep{palm2}, which is available via API calls.\footnote{\url{https://developers.generativeai.google/products/palm}} We use temperature 0.0 to increase factuality. An example of our \thedataset prompt with the preceding instructions and two retrieved demonstrations is provided in Table~\ref{tab:qsum_prompt}. We also include the respective QSum-S prompt in Table~\ref{tab:qsum_alce_prompt}, which includes \thedataset examples that were automatically converted  to have ALCE-style sentence-level citations instead of span-level quotes, and has the light ALCE~\citep{Gao2023EnablingLL} prompt instructions and formatting.

\subsection{In-context exemplar selection ablation}
We evaluate the effect of our Sentece-T5 nearest neighbor approach for selecting in-context examples. Table~\ref{tab:knn_ablation} shows that this approach provides better results compared to randomly selecting examples from the dataset. With random examples, the “fluency” (Rouge-L) drops a bit, and the “preciseness” and “comprehensiveness” scores drop more (by about 5 points each). This indicates that the random examples still help the model to generate relatively good answers, but the related questions are more helpful in guiding the model to extract the right spans from the sources. Interestingly the “random” 4-shot results are still higher than the “nearest neighbor” 1-shot (especially in comprehensiveness), suggesting that more \thedataset examples help the model align better with the SEMQA task, and that examples with more similar questions help the model extract better spans for the semi-extractive answer. 

Overall, given the diversity of questions in \thedataset, it is not surprising that different questions lead to different strategies for composing semi-extractive answers. Retrieving similar questions and their answers can help guide the model towards answering the examined question. This shows the value of our curated \thedataset dataset.

\begin{table}[ht]
    \centering
    \small
\resizebox{\linewidth}{!}{%
\begin{tabular}{ldddd}
\toprule
Exemplars &  \hd{Rouge-L} &  \hd{Sem-F1} &  \hd{Sem-Rec} &  \hd{SEMQA} \\
\midrule
Random             &      61.23 &         57.14 &   72.19 &     59.15 \\
KNN             &      62.71 &         62.11 &   77.32 &     62.41 \\

\bottomrule
\end{tabular}

}%

    \caption{Our nearest neighbor question selection from the \thedataset training set for the in-context few-shot prompt (\S\ref{sec:exp_eval}) leads to better performance than with randomly sampled examples. Reported results are with 4-shot PaLM2-Bison model.}
    \label{tab:knn_ablation}
\end{table}

\begin{table*}[t]
\centering
% \scriptsize
\footnotesize
% \begin{small}
% % \begin{minipage}{\linewidth}
        \resizebox{\linewidth}{!}{
\begin{tabular}{p{0.99\linewidth}}
\toprule
% \verbatim
% \begin{minipage}{\linewidth}
% \raggedright

% \begin{verbatim}
Answer the question by summarizing the given sources while explicitly copying spans from the sources. When copying a span, use brackets and the respective source number to indicate that this span was copied. Use explicit copying as much as possible and for all factual statements, while preserving fluency. Make sure to use all relevant sources and properly quote them. Here are some examples:\\ 
Question: how much power does a wind turbine produce?\\ 
 \ [1] Compact wind acceleration turbine: It is generally thought that since the amount of power produced by a wind turbine is proportional to the cube of the wind speed, any acceleration benefit is potentially statistically significant in the economics of wind. As noted though this is an inaccurate as it ignores the impact of the exit to area ratio and is therefore an apples to oranges comparison. In the case of a typical CWAT/DAWT the power result in perfect theoretical operation once adjusted for the area of the shroud is actually the square of the velocity at the rotor. As the CWAT/DAWT diverges from theoretical function the power increase drops significantly according\\ 
 \ [2] Sustainable architecture: roof ledge. Small-scale rooftop wind turbines have been known to be able to generate power from 10\% to up to 25\% of the electricity required of a regular domestic household dwelling. Turbines for residential scale use are usually between 7 feet (2 m) to 25 feet (8 m) in diameter and produce electricity at a rate of 900 watts to 10,000 watts at their tested wind speed. Building integrated wind turbine performance can be enhanced with the addition of an aerofoil wing on top of a roof mounted turbine. Solar water heaters, also called solar domestic hot water systems, can\\ 
 \ [3] Turby wind turbine: can because horizontal axis (HAWT) types cannot change their pitch to face the wind directly. The turbine measures 2.0m (6'7") in diameter by 2.9m (9'6") high (including generator), and weighs 136 kg (300 lb). It is specified to generate power in winds of between 4 m/s (9 mph, 7.8kts) and 14 m/s (31 mph, 27.2kts), and can survive winds of 55 m/s (123 mph, 107kts). The rated power at 14 m/s is 2.5 kW (3.35 hp). The AC output from the synchronous generator is rectified to DC, then inverted to AC at 230V 50 Hz. Core International developed the turbine\\ 
Quoted summary: One source states the [ 1 amount of power produced by a wind turbine is proportional to the cube of the wind speed ] . Other sources state [ 2 Turbines for residential scale use ] [ 2 produce electricity at a rate of 900 watts to 10,000 watts ] , and [ 3 is specified to generate power in winds of between 4 m/s (9 mph, 7.8kts) and 14 m/s (31 mph, 27.2kts) ] .\\ 
\\ 
Question: a component is what?\\ 
 \ [1] Modular programming: in Dart, Go or Java) is sometimes used instead of module. In other implementations, this is a distinct concept; in Python a package is a collection of modules, while in Java 9 the introduction of the new module concept (a collection of packages with enhanced access control) is planned. Furthermore, the term "package" has other uses in software (for example .NET NuGet packages). A component is a similar concept, but typically refers to a higher level; a component is a piece of a whole system, while a module is a piece of an individual program. The scale of the term\\ 
 \ [2] Physical body: the system at a point in time changes from identifying the object to not identifying it. Also an object's identity is created at the first point in time that the simplest model of the system consistent with perception identifies it. An object may be composed of components. A component is an object completely within the boundary of a containing object. In classical mechanics a physical body is collection of matter having properties including mass, velocity, momentum and energy. The matter exists in a volume of three-dimensional space. This space is its extension. Under Newtonian gravity the gravitational field further away\\ 
Quoted summary: [ 1 A component is a similar concept, but typically refers to a higher level; a component is a piece of a whole system, while a module is a piece of an individual program ] in terms of [ 1 Modular programming ] . Whereas in the [ 2 Physical body ] , a [ 2 component is an object completely within the boundary of a containing object ] .\\ 
\\ 
Question: how can energy efficiency be improved?\\ 
 \ [1] HVAC. forced air systems, which are now widely used in churches, schools and high-end residences, are A drawback is the installation cost, which can be slightly higher than traditional HVAC systems: Energy efficiency can be improved even more in central heating systems by introducing zoned heating. This allows a more granular application of heat, similar to non-central heating systems. Zones are controlled by multiple thermostats. In water heating systems the thermostats control zone valves, and in forced air systems they control zone dampers inside the vents which selectively block the flow of air. In this case, the control system is very\\ 
 \ [2] Electrical efficiency: is valid only for non-reactive source and load impedances. High efficiency is particularly relevant in systems that can operate from batteries. Inefficiency may require weighing the cost either of the wasted energy, or of the required power supply, against the cost of attaining greater efficiency. Efficiency can usually be improved by choosing different components or by redesigning the system. Inefficiency probably produces extra heat within the system, which must be removed if it is to remain within its operating temperature range. In a climate-controlled environment, like a home or office, heat generated by appliances may reduce heating costs or increase\\ 
 \ [3] Efficient energy use: Efficient energy use, sometimes simply called energy efficiency, is the goal to reduce the amount of energy required to provide products and services. For example, insulating a home allows a building to use less heating and cooling energy to achieve and maintain a comfortable temperature. Installing LED lighting, fluorescent lighting, or natural skylight windows reduces the amount of energy required to attain the same level of illumination compared to using traditional incandescent light bulbs. Improvements in energy efficiency are generally achieved by adopting a more efficient technology or production process or by application of commonly accepted methods\\ 
 \ [4] Nuclear power phase-out: non-renewable energy sources. Renewable energy encompasses wind, biomass (such as landfill gas and sewage gas), hydropower, solar power (thermal and photovoltaic), geothermal, and ocean power. These renewable sources serve as alternatives to conventional power generation from thermal power stations run on nuclear or fossil fuels. A significant part of energy transition is reducing consumption by energy conservation and improvements in energy efficiency, an example is improved insulation for buildings; or improved energy efficiency by cogeneration of heat and power. Smart meters are able to charge higher prices at the time consumption peaks during the day, thereby causing electricity demand to\\ 
Quoted summary:\\
\bottomrule
\end{tabular}
}%
% \end{minipage}

% \end{verbatim}
% \end{minipage}
% \vspace{-4pt}
\caption{Example 2-shot \thedataset prompt with two examples from the training set retrieved by question similarity. The demonstration summaries for each example are randomly picked out of the available references.}\label{tab:qsum_prompt}
% \end{small}
\end{table*}

\begin{table*}[t]
\centering
% \scriptsize
\footnotesize
% \begin{small}
% % \begin{minipage}{\linewidth}
        \resizebox{\linewidth}{!}{
\begin{tabular}{p{0.99\linewidth}}
\toprule
% \verbatim
% \begin{minipage}{\linewidth}
% \raggedright

% \begin{verbatim}
Instruction: Write a high-quality answer for the given question using only the provided search results and cite them properly using [1][2][3].\\ 
Question: how much power does a wind turbine produce?\\ 
 \ [1] Compact wind acceleration turbine: It is generally thought that since the amount of power produced by a wind turbine is proportional to the cube of the wind speed, any acceleration benefit is potentially statistically significant in the economics of wind. As noted though this is an inaccurate as it ignores the impact of the exit to area ratio and is therefore an apples to oranges comparison. In the case of a typical CWAT/DAWT the power result in perfect theoretical operation once adjusted for the area of the shroud is actually the square of the velocity at the rotor. As the CWAT/DAWT diverges from theoretical function the power increase drops significantly according\\ 
 \ [2] Sustainable architecture: roof ledge. Small-scale rooftop wind turbines have been known to be able to generate power from 10\% to up to 25\% of the electricity required of a regular domestic household dwelling. Turbines for residential scale use are usually between 7 feet (2 m) to 25 feet (8 m) in diameter and produce electricity at a rate of 900 watts to 10,000 watts at their tested wind speed. Building integrated wind turbine performance can be enhanced with the addition of an aerofoil wing on top of a roof mounted turbine. Solar water heaters, also called solar domestic hot water systems, can\\ 
 \ [3] Turby wind turbine: can because horizontal axis (HAWT) types cannot change their pitch to face the wind directly. The turbine measures 2.0m (6'7") in diameter by 2.9m (9'6") high (including generator), and weighs 136 kg (300 lb). It is specified to generate power in winds of between 4 m/s (9 mph, 7.8kts) and 14 m/s (31 mph, 27.2kts), and can survive winds of 55 m/s (123 mph, 107kts). The rated power at 14 m/s is 2.5 kW (3.35 hp). The AC output from the synchronous generator is rectified to DC, then inverted to AC at 230V 50 Hz. Core International developed the turbine\\ 
Answer: One source states the amount of power produced by a wind turbine is proportional to the cube of the wind speed [1]. Other sources state Turbines for residential scale use produce electricity at a rate of 900 watts to 10,000 watts , and is specified to generate power in winds of between 4 m/s (9 mph, 7.8kts) and 14 m/s (31 mph, 27.2kts) [2] [3].\\ 
\\ 
Question: a component is what?\\ 
 \ [1] Modular programming: in Dart, Go or Java) is sometimes used instead of module. In other implementations, this is a distinct concept; in Python a package is a collection of modules, while in Java 9 the introduction of the new module concept (a collection of packages with enhanced access control) is planned. Furthermore, the term "package" has other uses in software (for example .NET NuGet packages). A component is a similar concept, but typically refers to a higher level; a component is a piece of a whole system, while a module is a piece of an individual program. The scale of the term\\ 
 \ [2] Physical body: the system at a point in time changes from identifying the object to not identifying it. Also an object's identity is created at the first point in time that the simplest model of the system consistent with perception identifies it. An object may be composed of components. A component is an object completely within the boundary of a containing object. In classical mechanics a physical body is collection of matter having properties including mass, velocity, momentum and energy. The matter exists in a volume of three-dimensional space. This space is its extension. Under Newtonian gravity the gravitational field further away\\ 
Answer: A component is a similar concept, but typically refers to a higher level; a component is a piece of a whole system, while a module is a piece of an individual program in terms of Modular programming [1]. Whereas in the Physical body , a component is an object completely within the boundary of a containing object [2].\\ 
\\ 
Question: how can energy efficiency be improved?\\ 
 \ [1] HVAC. forced air systems, which are now widely used in churches, schools and high-end residences, are A drawback is the installation cost, which can be slightly higher than traditional HVAC systems: Energy efficiency can be improved even more in central heating systems by introducing zoned heating. This allows a more granular application of heat, similar to non-central heating systems. Zones are controlled by multiple thermostats. In water heating systems the thermostats control zone valves, and in forced air systems they control zone dampers inside the vents which selectively block the flow of air. In this case, the control system is very\\ 
 \ [2] Electrical efficiency: is valid only for non-reactive source and load impedances. High efficiency is particularly relevant in systems that can operate from batteries. Inefficiency may require weighing the cost either of the wasted energy, or of the required power supply, against the cost of attaining greater efficiency. Efficiency can usually be improved by choosing different components or by redesigning the system. Inefficiency probably produces extra heat within the system, which must be removed if it is to remain within its operating temperature range. In a climate-controlled environment, like a home or office, heat generated by appliances may reduce heating costs or increase\\ 
 \ [3] Efficient energy use: Efficient energy use, sometimes simply called energy efficiency, is the goal to reduce the amount of energy required to provide products and services. For example, insulating a home allows a building to use less heating and cooling energy to achieve and maintain a comfortable temperature. Installing LED lighting, fluorescent lighting, or natural skylight windows reduces the amount of energy required to attain the same level of illumination compared to using traditional incandescent light bulbs. Improvements in energy efficiency are generally achieved by adopting a more efficient technology or production process or by application of commonly accepted methods\\ 
 \ [4] Nuclear power phase-out: non-renewable energy sources. Renewable energy encompasses wind, biomass (such as landfill gas and sewage gas), hydropower, solar power (thermal and photovoltaic), geothermal, and ocean power. These renewable sources serve as alternatives to conventional power generation from thermal power stations run on nuclear or fossil fuels. A significant part of energy transition is reducing consumption by energy conservation and improvements in energy efficiency, an example is improved insulation for buildings; or improved energy efficiency by cogeneration of heat and power. Smart meters are able to charge higher prices at the time consumption peaks during the day, thereby causing electricity demand to\\ 
Answer:\\
\bottomrule
\end{tabular}
}%
% \end{minipage}

% \end{verbatim}
% \end{minipage}
% \vspace{-4pt}
\caption{An example 2-shot QSum-S prompt that converts \thedataset examples to ALCE-style sentence-level citations, and also provides the ALCE instructions. This question is similar to the one in Table~\ref{tab:qsum_prompt} showing the matching Qsum prompt for reference.}\label{tab:qsum_alce_prompt}
% \end{small}
\end{table*}

\section{Additional Results}
\label{app:more_results}

Table~\ref{tab:example_answer_large} shows an example answer of PaLM2 Unicorn 5-shot model with a SEMQA \thedataset-based prompt, and a PaLM2 Bison with ALCE prompt answer to the same question.

\begin{table*}[t]
\centering
\begin{small}
% \begin{minipage}{\linewidth}
        \resizebox{\linewidth}{!}{
% \begin{tabular}{lp{0.15\linewidth}p{0.4\linewidth}p{0.08\linewidth}p{0.12\linewidth}r}
\begin{tabular}{p{0.1\linewidth}|p{0.90\linewidth}}
\toprule
Question & \textit{where is st mark's church located?} \\
\midrule
% \multicolumn{2}{l}{\emph{...:}} \\
%  \tcbox[colback=hl_1]
Sources & [1] Church of St: Luke and The Epiphany (Philadelphia). southwestern edge of development within the original boundaries of Philadelphia (it would take almost another 10 years before an Episcopal church was organized across Broad Street and below Chestnut Street – \tcbox[colback=hl_1]{St. Mark’s Church} \tcbox[colback=hl_1]{at 16th and Locust Streets}). These men advanced the money to build the church and were repaid through the sale of pews. Several sites were considered including the corner Broad and Locust Streets and the corner of 13th and Spruce Streets. These locations were rejected due to cost. The committee chose the present site of the church and selected Philadelphia architect Thomas S. Stewart to design \\
& [2] Charles Eamer Kempe: or wheatsheaf, taken from Kempe's own coat of arms. The mid-Victorian period were important years in the history of the design of English churches and Kempe’s influence is found in numerous examples, many in his home county of Sussex which has 116 examples of his work. The works at \tcbox[colback=hl_2]{St Mark’s, Staplefield} \tcbox[colback=hl_2]{near Horsham, West Sussex} dating from 1869 are regarded as especially important, representing the earliest of three known examples of Kempe’s wall painting. They contain key elements of Kempe’s figurative work. The angels holding the scroll are magnificently apparelled and the borders of their cloaks are embellished with \\
& [3] St: Mark's Cathedral, Bangalore. \tcbox[colback=hl_3]{St. Mark's Cathedral} is named after Saint Mark (believed to be the first gospel writer) and is located \tcbox[colback=hl_3]{at the west end of Mahatma Gandhi Road, MacIver Town, Bangalore}. Its architecture is inspired by the 17th century St Paul's Cathedral. The church serves as the Cathedral (Ecclesia Matar) of the Church of South India, Karnataka Central Diocese. Found in 1808, the cathedral celebrated its 200 years bicentenary in 2007-8. The cathedral is open for all people irrespective of religion or faith. The large grounds of the cathedral with numbers a trees offers a peaceful refugee from \\
& [4] \tcbox[colback=hl_4]{Saint Mark's Coptic Orthodox Church} (London): \tcbox[colback=hl_4]{Saint Mark's Coptic Orthodox Church} is a church \tcbox[colback=hl_4]{in Kensington, London}, located at the south-west corner of the intersection of Allen Street and Scarsdale Villas. It is dedicated to the founder of the Church of Alexandria, Mark the Evangelist. The building was previously Saint John's Presbyterian Church and was opened in 1863. Purchased from the Scottish Presbyterian church in 1975, Saint Mark's was the first Coptic Orthodox church in Europe. It was the seed church of the Coptic Orthodox Church in Britain and Ireland, first served by three hieromonks who later became members of\\
\midrule
PaLM2 Unicorn 5-shot SEMQA prompt & Sources state \tcbox[colback=hl_1]{St. Mark’s Church} is \tcbox[colback=hl_1]{at 16th and Locust Streets} , \tcbox[colback=hl_2]{St Mark’s, Staplefield} is \tcbox[colback=hl_2]{near Horsham,} \tcbox[colback=hl_2]{West Sussex} , \tcbox[colback=hl_3]{St. Mark's Cathedral} is \tcbox[colback=hl_3]{at the west end of Mahatma Gandhi Road, MacIver Town, Bangalore} , and \tcbox[colback=hl_4]{Saint Mark's Coptic Orthodox Church} is \tcbox[colback=hl_4]{in Kensington, London} .\\
\midrule
PaLM2 Bison ALCE prompt & St. Mark's Church is located at the west end of Mahatma Gandhi Road, MacIver Town, Bangalore [3].\\
\bottomrule
\end{tabular}
% \end{minipage}
}%
\vspace{-4pt}
\caption{Example generated answers with a 5-shot PaLM2 Unicorn model using our SEMQA prompt with \thedataset examples, and a 4-shot PaLM2 Bison using the ALCE prompt. Colored spans show the spans that the SEMQA model marked as explicit extractions from the respective source, by color. In this example, the ALCE output answered only based on source number 3. The SEMQA answer covers all 4 passages with in-line quotes to attribute the respective sources.}\label{tab:example_answer_large}
\end{small}
\end{table*}

\end{document}